\documentclass[10pt, a4paper, copyright, logo, nonumbering]{instadeep}

\usepackage[utf8]{inputenc} 
\usepackage[T1]{fontenc}
\DeclareUnicodeCharacter{2032}{'}

\usepackage{comment}
\usepackage{adjustbox}
\usepackage{tabularx}
\usepackage{makecell}
\usepackage{tabulary}
\usepackage{lscape}
\usepackage{minted}

\usepackage{caption}
\usepackage{hyperref}
\usepackage{titletoc}

\title{InstaGeo: Compute-Efficient Geospatial Machine Learning from Data to Deployment}
\headertitle{Scalable End-to-End Geospatial Machine Learning Framework}

\correspondingauthor{i.yusuf@instadeep.com, a.pretorius@instadeep.com}

\renewcommand{\today} 

\author[**,1]{Ibrahim Salihu Yusuf}
\author[1]{Iffanice Houndayi}
\author[1]{Rym Oualha}
\author[1]{Mohamed Aziz Cherif}
\author[1]{Kobby Panford-Quainoo}
\author[**,1]{Arnu Pretorius}

\affil[1]{InstaDeep}
\affil[**]{Corresponding authors: i.yusuf@instadeep.com, a.pretorius@instadeep.com}

\newcommand{\brandname}{InstaGeo\xspace}

\usepackage{xspace}
\usepackage{amsmath}
\usepackage{amsfonts}       

\definecolor{chineseblue}{rgb}{0.21,0.31,0.58}
\definecolor{myred}{rgb}{0.8,0,0}
\definecolor{mygreen}{rgb}{0,0.6,0}
\definecolor{myblue}{rgb}{0,0,0.7}
\definecolor{stanfordblue}{HTML}{006eb8}
\definecolor{amethyst}{rgb}{0.6, 0.4, 0.8}
\definecolor{arsenic}{rgb}{0.23, 0.27, 0.29}

\definecolor{myblue2}{rgb}{0,0,0.6}

\usepackage{caption}

\begin{abstract}
Open access multispectral imagery from Landsat 8–9, Sentinel-2 and similar missions has spurred the rise of geospatial foundation models (GFMs) fine-tuned for various downstream humanitarian and societal tasks. However, the deployment of these models faces two major obstacles: (i) the lack of dedicated geospatial data pipelines and (ii) the prohibitive size of fine-tuned models. These challenges arise because published GFMs currently do not include pipelines for generating input data from raw satellite imagery, and models derived for each downstream task retain the architecture and complexity of the pre-trained GFM encoder. To address these challenges, we introduce \textbf{\brandname}, an open-source, end-to-end geospatial machine learning framework that combines automated data curation for converting raw satellite imagery into a model-ready format; a model component with task-specific distillation for transforming large GFMs into compute-efficient models whose complexity matches task difficulty; and a component for deploying models as interactive web-map application for operational use. Using \brandname, we faithfully reproduced the datasets that underlie three published studies from scratch and trained models with performance differences of \(- 0.73\) percentage point (pp) mean intersection-over-union (mIoU) for flood mapping, \(- 0.20\) pp mIoU for multitemporal crop segmentation, and \(+ 1.79\) pp mIoU for desert locust breeding ground prediction. Our task-specific distilled models are compute efficient and up to \(8\times\) smaller than those obtained via standard fine-tuning, significantly reducing inference FLOPs and thereby \(\mathrm{CO_{2}}\) emissions with minimal performance degradation. Due to the ease of use of \brandname's data pipeline, we curated a larger crop segmentation dataset, achieving a new state-of-the-art mIoU of \textbf{60.65 \%}, an improvement of \textbf{12 pp} over the previous baseline. Finally, we showcase how \brandname can significantly accelerate the data-to-deployment cycle, allowing a user to move from data preparation to model deployment within a single working day. By unifying data preparation, model compression, and visual analysis within a single open-source framework, \brandname transforms research-grade GFMs into practical, low-carbon tools suitable for real-time, large-scale environmental monitoring. This unified approach has the potential to shift Earth Observation (EO) research from model-centric competition toward data-quality and application-centric innovation. We release the source code of \brandname, along with datasets, model checkpoints, and bash scripts used for data curation and model training, at \url{https://github.com/instadeepai/InstaGeo-E2E-Geospatial-ML.git}.
\end{abstract}

\begin{document}

\maketitle

\newpage
\begin{figure}[h!]
\centering
\begin{center}
\includegraphics[width=1.0\textwidth]{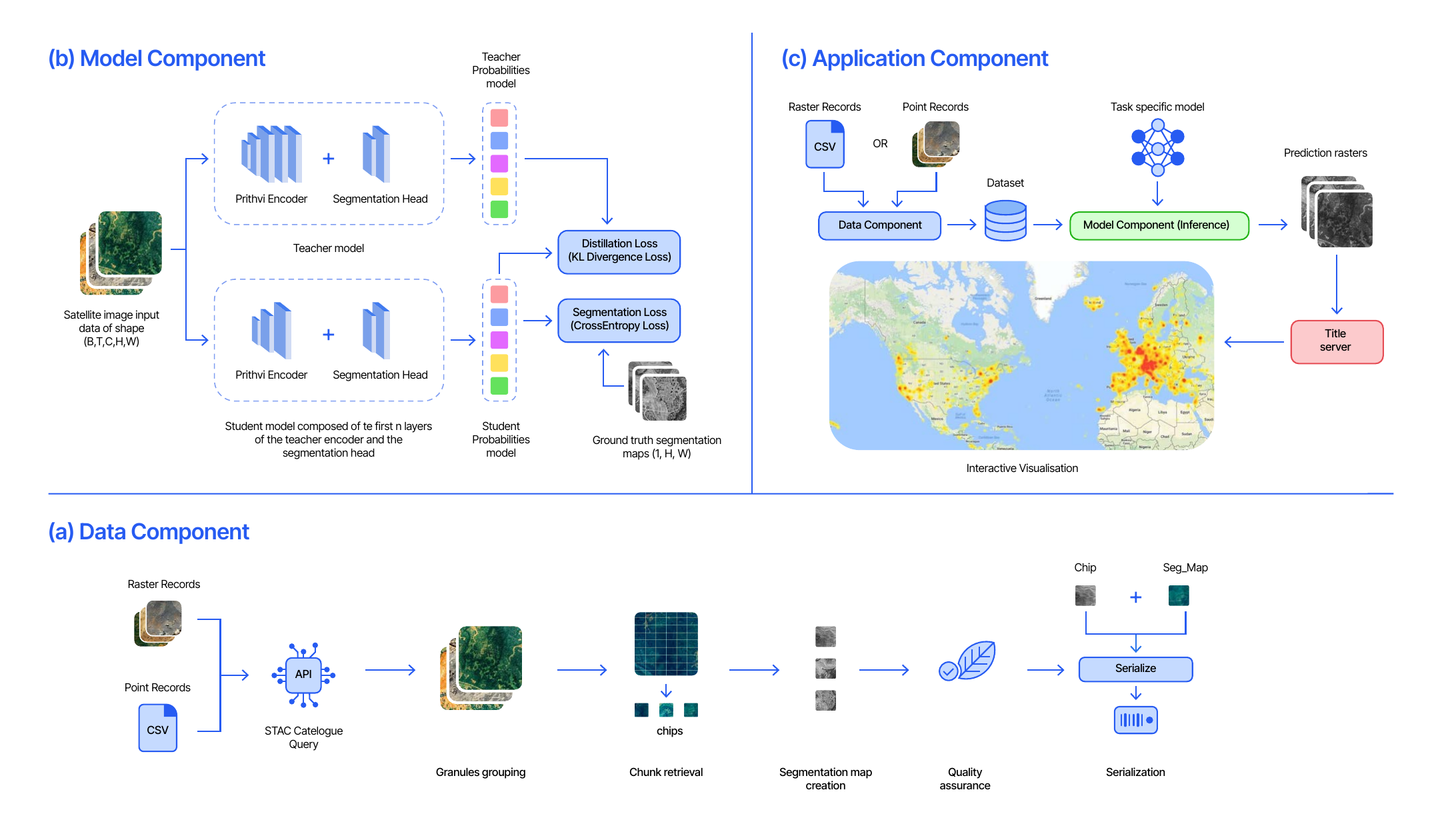}
\end{center}
    \caption{
    \textbf{Overview of \textit{InstaGeo} for geospatial machine learning.}  
    (a)~\emph{Data component.}  Point or raster observations are ingested, matched to satellite granules via STAC API search, grouped by common acquisition, cut into multitemporal “chips”, with labels rasterised into segmentation maps, cloud‑masked and serialised as GeoTIFF pairs.  
    (b)~\emph{Model component.}  A frozen GFM encoder plus segmentation head acts as the teacher; a shallower student shares the first \(n\) encoder layers.  The student is trained with a combined segmentation loss (ground‑truth maps) and distillation loss (KL divergence to the teacher logits), yielding a lightweight task‑specific model.  
    (c)~\emph{Application component.}  The trained model runs inference on new sampled images to generate prediction rasters, which are then served via a tile server using TiTiler and rendered as interactive web‑map layers.  Together, these three components convert raw observations into operational geospatial intelligence in a single, reproducible workflow.
}
    \label{fig:panel1}
\end{figure}

Open-access multispectral missions, such as NASA/USGS Landsat 8–9 \cite{roy2014landsat8, masek2020landsat9} and ESA Sentinel-2A/B \cite{drusch2012sentinel2, gascon2014calval}, have reshaped Earth Observation (EO) research, enabling applications that range from disaster response and ecosystem monitoring to crop yield forecasting, land use assessment and deforestation tracking \cite{kansakar2016review, phiri2020sentinel2review}. The Harmonized Landsat Sentinel-2 archive (HLS) \cite{HLS} extends this impact by providing 30m imagery every 2–3 days, yielding an unrivalled open-access spatio-temporal record of the planet.  
To capitalise on such data at scale, the community has turned to \emph{geospatial foundation models} (GFMs): large, self-supervised networks pre-trained on a large collection of satellite imagery and subsequently fine-tuned for various downstream tasks \cite{Tuia_2025}.  
Despite the progress achieved so far, two systemic obstacles still limit their operational use: the lack of (i) an open-source geospatial data pipeline that converts raw satellite imagery into a model-ready format and (ii) a scalable geospatial model deployment framework.

First, we consider the issue of geospatial data pipelines. To date, and to the best of our knowledge, \emph{no publicly released GFM\,\cite{satmae,dofa,gfm,jakubik2023foundationmodelsgeneralistgeospatial,spectral_gpt,szwarcman2025prithvieo20,spradlin2024satvisiontoa} includes an open-source geospatial data pipeline}; practitioners still have to convert raw satellite imagery into the exact chip tensors used during pre-training before they are able to use the GFM, an error-prone, complex and time-consuming task.  
Recent models illustrate the gap. \textbf{SatMAE} \cite{satmae} demonstrates masked-autoencoder pre-training on Sentinel-2 time series but ships only a 307M-parameter checkpoint, leaving users to build their own data pipeline. Teacher–student frameworks that ingest non-temporal stacks such as \textbf{GFM} \cite{gfm} and \textbf{DOFA} \cite{dofa} reduce training cost and CO\(_2\) emissions, yet their distilled models (300M parameters) also assume that the input tensors have already been carved from raw satellite imagery. Architectures that ingest temporal stacks and preserve full 3-D spatial–spectral structure, e.g.\ \textbf{SpectralGPT} \cite{spectral_gpt} (100–600M parameters), further raise the deployment bar while still omitting preprocessing code. Even large-scale HLS pre-training, \textbf{Prithvi-EO-2.0} \cite{szwarcman2025prithvieo20}, attains state-of-the-art GEO-Bench \cite{lacoste2023geobench} scores only at a 600M-parameter footprint and does not provide tooling for converting satellite image tiles into the chips used during training. In every case, the burden of dataset construction falls on the practitioner and remains a major barrier to GFM adoption for operational use.

Second, the standard practice of using GFMs involves replacing the pre-trained decoder with a custom head and then fine-tuning this modified model on a new downstream task, a procedure we refer to as \emph{vanilla fine-tuning}. This results in a monolithic model that maintains the same architecture and complexity regardless of the difficulty of the downstream task. However, downstream tasks vary widely in difficulty, ranging from simple to highly challenging. Consequently, this practice leads to excessive computational costs and carbon footprints, irrespective of the difficulty of the task. Moreover, retaining the fine-tuned model as-is typically results in a research artefact that rarely progresses to operational use, primarily due to the absence of open-source and user-friendly frameworks for deploying geospatial models. 

To address the above limitations, we present \textbf{\brandname}, a compute-efficient, end-to-end, open-source geospatial machine learning framework complete with a data pipeline, model training and inference infrastructure, and an application component for operational use. Our integrated geospatial data pipeline seamlessly converts geolocated point or polygon observations into a dataset of model-ready, cloud-masked HLS chips, based on user-defined spatial and temporal configurations. The model component transforms large pre-trained GFMs into specialized models for each downstream task through either vanilla fine-tuning or task-specific distillation. For task-specific distillation, we employ a teacher-student paradigm to derive a student model with fewer encoder layers while closely matching the teacher's performance. This results in lightweight student models whose complexity aligns with task difficulty, significantly reducing the number of parameters by up to \(8\times\) with minimal performance degradation. Lastly, a built-in web application serves as a portal to select areas of interest, trigger model inference over selected areas, render model predictions as interactive map layers, and export results as analytic overviews enabling non-technical users to easily obtain actionable insights. \brandname reduces the time from labeled data (point or field observations) to an operational model to one working day, which is critical for time-sensitive applications.

Across different downstream tasks \brandname student models rival vanilla fine-tuned teachers while using far fewer parameters and FLOPs, thereby lowering inference cost and the associated $\mathrm{CO_{2}}$ emissions. By unifying data preparation, model compression and visual analytics in a single open-source platform, InstaGeo transforms research-grade GFMs into practical, low-carbon tools for real-time, large-scale environmental monitoring. This work has the potential to move EO research away from model-centric benchmarks and toward innovation driven by data quality and real-world applications. To facilitate this, we provide the full \brandname\ source code, together with datasets, model checkpoints, and the accompanying scripts for data processing and training at \url{https://github.com/instadeepai/InstaGeo-E2E-Geospatial-ML.git}.
\section*{Results}

\subsection{Reproducing Published Benchmarks using \brandname's Data Pipeline}
\label{sec:replica-exp}
A key requirement to drive the broad adoption of GFMs is having a geospatial data pipeline that allows the deployment of both existing and newly fine‑tuned geospatial models. To facilitate the deployment of existing models, the data pipeline must accurately reproduce the exact chip tensors used during training, ensuring that the original model performance is preserved at inference time. Currently, no such open-source data pipeline exists. Given that \brandname provides such a pipeline, we therefore investigated whether it can faithfully \emph{replicate} the data pipelines underpinning three published studies on flood mapping \cite{szwarcman2025prithvieo20}, multi‑temporal crop classification \cite{jakubik2023foundationmodelsgeneralistgeospatial,szwarcman2025prithvieo20}, and locust breeding ground prediction \cite{yusuf2024geospatialapproachpredictingdesert}, while preserving model performance.

\begin{figure}[h!]
\begin{center}
\includegraphics[width=1.0\textwidth]{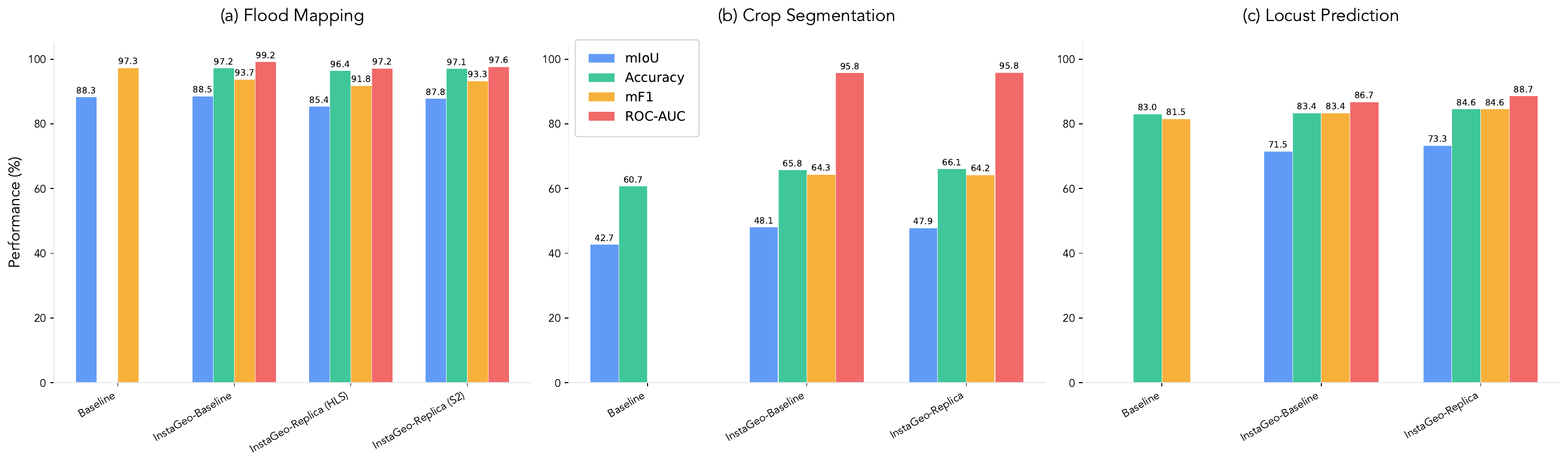}
\end{center}
\caption{ 
\textbf{\brandname reproduces unpublished data pipelines for published datasets.} For each task we the following performances: (i) the \emph{Baseline} reported in the original study, (ii) the \emph{\brandname-Baseline} trained using the authors’ original dataset but using \brandname's model component, and (iii) the \emph{\brandname-Replica} trained on a dataset entirely reconstructed using \brandname. \textbf{(a)} Performance comparison for the flood mapping task. The original dataset was created using Sentinel-2 imagery, and we created two replicas, one from HLS and another from Sentinel-2. \textbf{b} Performance comparison for the multi-temporal crop segmentation task. \textbf{(c)} Performance comparison for the desert locust breeding ground prediction task.
}
\label{fig:panel2}
\end{figure}

For each task, we used \brandname to build a replica of the original dataset that matches the spatial, temporal, and spectral specifications of the original study. We then re‑trained the corresponding GFM on this replica and compared its performance to that obtained when training on the dataset of the original authors. Figure~\ref{fig:panel2} (see also Table~\ref{table:replica_exp}) shows a comparison of the resulting scores. Across all tasks, the new models trained on our replica datasets reproduce the original performance within \(\pm 2\) percentage-point (pp) mIoU (-0.73 pp for flood mapping, -0.20 pp for crop classification, +1.79 pp for locust prediction)--differences minimal enough to reasonably attribute them to floating point precision. 
For flood mapping specifically, we produced two replicas: one from HLS and one from native Sentinel-2 tiles. The Sentinel-2 replica closely matches the baseline (within -0.73 pp mIoU), while the HLS replica trails by approximately $\sim\!3$pp. This performance shortfall can be attributed to the spectral adjustment made during HLS preprocessing, which modifies Sentinel-2 bands to align with the Landsat sensor \cite{HLS}.  These results demonstrate that, when the source imagery remains consistent, \brandname effectively replicates complex EO pipelines with minimal performance differences, ensuring reliable model deployment. Conversely, when the source imagery changes, the framework accurately highlights performance shifts arising from sensor differences. The bash scripts used for data preparation and model training are included in \brandname's open-source repository.

\begin{table}[ht]
\caption{ \textit{\brandname reproduces unpublished data pipelines.} For each task we report the performance of (i) the \emph{Baseline} performance reported in the corresponding study, (ii) the \emph{\brandname-Baseline} model trained on the authors’ data but using \brandname's model component, and (iii) the \emph{\brandname-Replica} model trained on a dataset reconstructed entirely with \brandname. The flood mapping replica has a version derived from HLS and another derived from Sentinel-2.
}
\label{table:replica_exp}
\centering
\resizebox{0.95\columnwidth}{!}{%
\begin{tabular}{l|cccccc}
\toprule
\textbf{Task} & \textbf{Model} & \textbf{GFM} & \textbf{mIoU (std)} & \textbf{Acc} & \textbf{mF1 (std)} & \textbf{ROC-AUC (std)} \\

\midrule
\multirowcell{3}{Flood Mapping} 
  & Baseline          & Prithvi-V1-100M & 88.3 (0.3) & --          & 97.3 (0.1) & -- \\
  & \brandname-Baseline & Prithvi-V1-100M & 88.53       & 97.24       & 93.71 & 99.16 \\
  & \brandname-Replica (HLS)  & Prithvi-V1-100M & 85.40      & 96.39       & 91.78 & 97.15 \\
  & \brandname-Replica (S2)  & Prithvi-V1-100M & 87.80      & 97.07       & 93.26 & 97.61 \\
\midrule
\multirowcell{3}{Multi-Temporal Crop \\ Segmentation (US)} 
  & Baseline          & Prithvi-V1-100M & 42.7       & 60.7        & --  & -- \\
  & \brandname-Baseline & Prithvi-V1-100M & 48.07       & 65.77       & 64.34 & 95.79 \\
  & \brandname-Replica  & Prithvi-V1-100M & 47.87      & 66.10       & 64.19 & 95.82 \\
\midrule
\multirowcell{3}{Locust Breeding \\ Ground Prediction} 
  & Baseline          & Prithvi-V1-100M & --      & 83.03          & 81.53 & -- \\
  & \brandname-Baseline & Prithvi-V1-100M & 71.51         & 83.39          & 83.39 & 86.74 \\
  & \brandname-Replica  & Prithvi-V1-100M & 73.30         & 84.60          & 84.60 & 88.66 \\
\bottomrule

\end{tabular}%
}
\end{table}

\subsection{Task‑Specific Distillation Produces Lightweight Models with Comparable Accuracy}
\label{sec:task_specific}
\begin{figure}[h!]
\begin{center}
\includegraphics[width=1.0\textwidth]{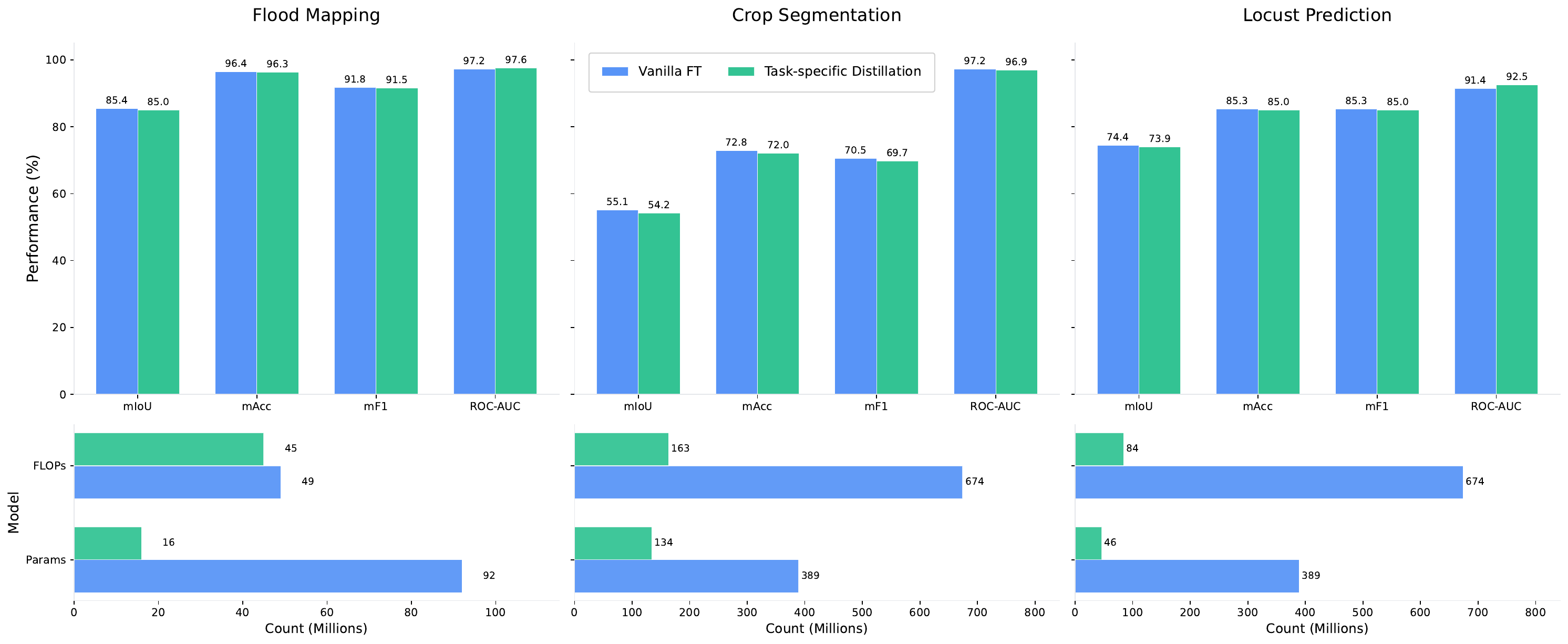}
\end{center}
\caption{
\textbf{Task-specific distillation yields lightweight models with comparable performance and reduced computational cost.} 
Each column corresponds to a downstream task, where the top row compares the performance of a vanilla fine-tuned model (\emph{Vanilla FT}) against a lightweight student model obtained via task-specific distillation across four evaluation metrics: mIoU, mAcc, mF1, and ROC-AUC. 
The bottom row illustrates the model characteristics by comparing the parameter counts (in millions) and estimated compute requirements (in GFLOPs) of the two models. 
Despite significantly reduced model size and compute, task-specific student models retain comparable performance, demonstrating the effectiveness of task-specific distillation.
}
\label{fig:panel3}
\end{figure}

 Our preliminary experiments revealed that, for some downstream tasks, small models trained from scratch can achieve decent performance. This observation suggests that inference costs and associated $CO_{2}$ emissions can be significantly reduced by matching the size of the fine-tuned model to the complexity of the task. To test this hypothesis, we adopt a task-specific distillation strategy (see \hyperref[sec:model-component]{Methods}), in which the student model retains only the number of encoder layers necessary for the task complexity. A vanilla fine-tuned model with frozen weights serves as the teacher, while the lightweight student model is trained to minimize the sum of a task-specific loss and a distillation loss computed between the logits of the student and teacher models. We apply this task-specific distillation approach separately for each task evaluated in our \hyperref[sec:replica-exp]{replica experiments}. Figure~\ref{fig:panel3} (see also Table~\ref{table:finetune-exp}) compares the vanilla fine-tuned teacher with the corresponding student derived using task‑specific distillation. Across all three benchmarks, the distilled models either recover or show a minor performance drop when compared to vanilla fine-tuning while reducing parameter count by almost an order of magnitude (5x for Flood Mapping, 2x for crop classification and 8x for locust breeding ground prediction). These results demonstrate that \brandname can reduce computational complexity and, hence, inference latency and energy consumption throughout the lifecycle of the model in deployment, thus lowering operational costs and facilitating wider adoption of GFMs.

 \begin{table} [ht]
\caption{\textbf{Task‑specific distillation preserves comparable performance while significantly reducing model size.}
For each downstream task, we report the performance of (i) \emph{Vanilla Fine‑tuning}, which updates all weights of the GFM and segmentation head, and (ii) \emph{Task‑specific Fine‑tuning}, where a lightweight student model is distilled from the same teacher. The encoder layer in the students is initialized from those of the teacher. Parameter counts for each model is reported and in all cases they are at least 2× smaller than the teacher, yielding significant inference‑time savings.} 
\label{table:finetune-exp} 
\centering 
\resizebox{0.95\columnwidth}{!}{%
\begin{tabular}{l|lccccccc} 

\toprule 

\textbf{Task} & \textbf{Method} & \textbf{GFM} & \textbf{Parameters} & \textbf{FLOPs} & \textbf{mIoU} & \textbf{mAcc} & \textbf{mF1 (std)} & \textbf{ROC-AUC (std)} \\

\midrule 

\multirowcell{2}{Flood Mapping} 
    & Vanilla FT & Prithvi-V1-100M & 92M & 49 & 85.40      & 96.39       & 91.78 & 97.15 \\
    & Task‑specific (student) & - & 16M & 45 & 84.99 & 96.31 & 91.51 & 97.60\\ 
    
\midrule 

\multirowcell{2}{Multi‑Temporal Crop \\ Segmentation (US)} 
    & Vanilla FT & Prithvi-V2-300M & 389M & 674  & 55.10      & 72.81       & 70.51 & 97.19 \\ 
    & Task‑specific (student) & - & 134M & 163 & 54.21 & 72.04 & 69.74 & 96.93 \\ 
    
\midrule 

\multirowcell{2}{Locust Breeding \\ Ground Prediction} 
    & Vanilla FT & Prithvi-V2-300M & 389M & 674 & 74.40 & 85.32 & 85.32 & 91.42 \\
    & Task‑specific (student)& - & 46M & 84 & 73.90 & 84.99 & 84.99 & 92.46 \\ 

\bottomrule 

\end{tabular}%
 }
\end{table}

\subsection{Task-level evaluation on expanded crop segmentation benchmark}
\label{sec:downstream_results}
\begin{figure}[h!]
\begin{center}
\includegraphics[width=0.7\textwidth]{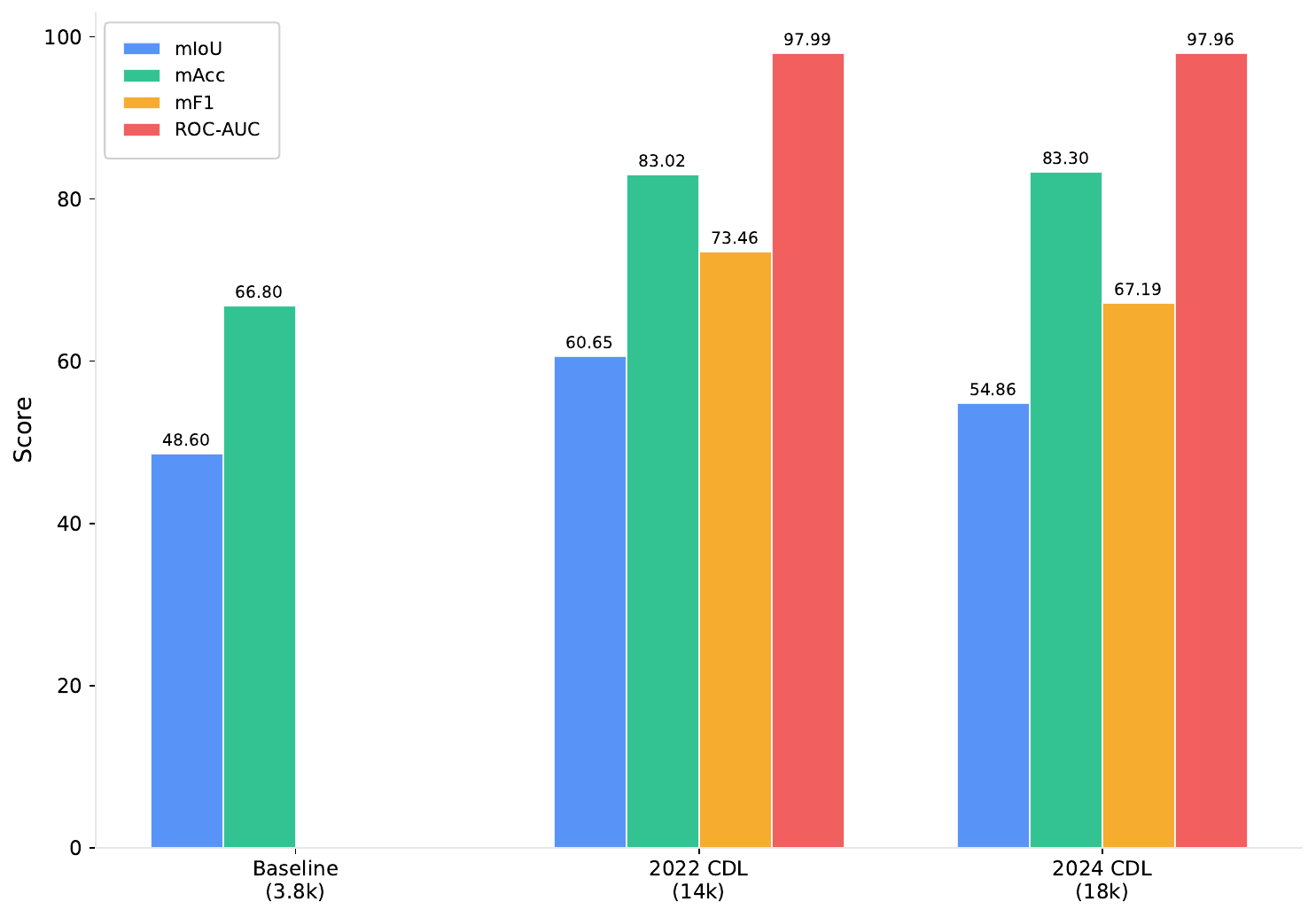}
\end{center}
\caption{
\textbf{Impact of larger CDL variants on multitemporal crop segmentation, showing mIoU, mAcc, mF1 and ROC–AUC for the published baseline (3.8k chips), expanded 2022 CDL (14k chips) and 2024 CDL (18k chips)}
}
\label{fig:panel4}
\end{figure}

Having demonstrated that \brandname\ can faithfully reproduce unpublished geospatial pipelines, we next evaluated its ability to enhance the multi-temporal crop segmentation benchmark. We began with the original Prithvi-EO-1.0 \cite{jakubik2023foundationmodelsgeneralistgeospatial} dataset of 3,854 chips derived from the 2022 U.S. Cropland Data Layer (CDL)\footnote{\url{https://croplandcros.scinet.usda.gov/}}. The dataset was divided into the 80:20 ratio between training and validation, with performance metrics reported on the 771-chip validation split. We regenerated and expanded this to 14,061 chips, divided into training, validation, and test sets in the 70:10:20 ratio, and fine-tuned Prithvi-V2-300M under the vanilla fine-tuning scheme on the 2022 U.S CDL data. Despite reporting metrics on our test set, we observed a 12pp increase in mIoU and a 16pp increase in global accuracy over the original baseline (Figure~\ref{fig:panel4}; Table~\ref{table:new-us-cdl-exp}), establishing a new benchmark performance. Using the newly released 2024 CDL, processed via an updated workflow \cite{li2024cloudagriculture}, we generated a dataset with 18,675 chips and fine-tuned under identical settings. Although the 16pp accuracy gain persisted, the improvement in mIoU decreased to 6 pp, likely reflecting the revised label generation pipeline.

\begin{table}[ht]
\caption{\textbf{Impact of larger CDL variants on multitemporal crop segmentation.}  
Prithvi-V2-300M was fine-tuned on three datasets: the published baseline (3,854 chips), an expanded 2022 CDL variant (14,061 chips) and a 2024 CDL variant (18,675 chips).  
Bold numbers denote the best value in each column.}
\label{table:new-us-cdl-exp}
\centering
\resizebox{0.9\columnwidth}{!}{%
\begin{tabular}{l|cccc}
\toprule
\textbf{Dataset variant} & \textbf{mIoU} & \textbf{mAcc} & \textbf{mF1} & \textbf{ROC–AUC} \\
\midrule
Published baseline (Prithvi, 2022, 3.8k) & 48.60 & 66.80 & - & - \\
Expanded 2022 CDL (Ours, 14k)       & \textbf{60.65} & 83.02 & \textbf{73.46} & \textbf{97.99} \\
2024 CDL (Ours, 18k)                & 54.86 & \textbf{83.30} & 67.19 & 97.96 \\
\bottomrule
\end{tabular}}%
\end{table}

\subsection{\brandname shortens the data-to-deployment cycle}
Conventional EO workflows often hinge on a patchwork of specialized tools; one for data download, another for preprocessing, a separate package for model training, and yet another solution for serving results, each introducing its own steep learning curve and potential bottleneck. \brandname breaks down these silos by unifying every stage into a single end-to-end framework, thereby flattening the overall learning trajectory and slashing the time required to go from raw satellite imagery to an operational inference service. To demonstrate these end‑to‑end gains, we applied InstaGeo to the flood‑mapping task and tracked the performance across data preparation, model development, and deployment.

Using the chip\_creator module, we extracted 424 HLS chips ($256 \times 256$ pixels) for training, validation, and test splits in just 1h 43min on a standard 8‑core workstation with 32GB of RAM. We then fine‑tuned Prithvi‑V1‑100M for 100 epochs (batch\_size=16), completing training in 3h 14min and producing a 92M‑parameter model that required approximately 49.3 GFLOPs of compute. To further boost model efficiency, we distilled the 92M parameter teacher into a smaller 14M parameter student in an additional 3h 10min, cutting compute requirements nearly in half to 27.5 GFLOPs, with negligible performance loss.

Deployment through the application component, which comprises an API and at least three workers (for data extraction, models inference and visualization preparation) on the backend ; and an interactive web-map application on the frontend, took only 30 minutes. Together, the total wall‑clock time from raw label ingestion to a fully functional map‑based inference service was only 8h 37min. Such rapid turnaround is critical for time sensitive applications (for example, emergency flood response), and highlights \brandname’s ability to transform EO research assets into operational tools within a single working day.
\section*{Discussion}
This study presents \textit{\brandname}, an open-source, end-to-end framework that converts raw satellite imagery and \emph{in situ} labels into operational, compute-efficient geospatial models. At its core lies a fully automated data pipeline that transforms raw imagery into task-ready, multi-temporal chips. A flexible model component then fine-tunes models for each downstream task, while another component embeds them into a built-in interactive web-map application. To our knowledge, \brandname is the first open-source solution that seamlessly guides users from \emph{in situ} labels through model development to operational deployment without switching tools.

We showed that \brandname data pipeline (\emph{chip\_creator}) can both reconstruct the datasets that underpin existing GFMs as well as generate entirely new ones. We rebuilt three published benchmarks from scratch since the original studies did not publish their data pipeline, which is required for operationalising the model. When the original data source (HLS, Sentinel-2, etc.) is retained, the model derived using our replica dataset recover the performance of the original study: multi-temporal crop classification \cite{jakubik2023foundationmodelsgeneralistgeospatial,szwarcman2025prithvieo20}, locust breeding ground prediction \cite{yusuf2024geospatialapproachpredictingdesert} and Sentinel-2 flood mapping sets \cite{szwarcman2025prithvieo20} differ only by –0.20 pp, +1.79 pp and 0.73 pp mIoU, respectively (See Figure ~\ref{fig:panel2} and Supplementary Table \ref{table:replica_exp}). Given that \emph{chip\_creator} supports HLS and Sentinel-2 satellite data products, and the original flood mapping dataset was derived using Sentinel-2, a parallel replica derived from HLS imagery was created that scored \(\sim\!3\) pp lower than the baseline, a gap that can be attributed to the spectral adjustment applied in HLS preprocessing \cite{HLS}. These experiments demonstrate that InstaGeo lets practitioners operationalise both legacy and newly fine-tuned GFMs with negligible performance loss and minimal effort, while its open-source, reproducible pipeline can serve as a common foundation for the EO community.

For many applications, particularly those that require frequent inference over large areas, the computational demands of a full-scale GFM can drive costs to be prohibitively high and exclude users without GPU resources. Our vision for \brandname is to allow users with reasonably equipped computers to use open-source satellite data to build geospatial solutions. Using our teacher–student distillation strategy, we trimmed vanilla fine-tuned teachers by up to 8× in parameters and 2× in FLOPs, yet the distilled students matched their teachers to within a few-tenths of a point in mean IoU, accuracy and F1 (see Figure ~\ref{fig:panel3} and Table ~\ref{table:finetune-exp}). The result reinforces a key principle: When model capacity is aligned with task complexity, much smaller, task-specific models can deliver teacher-level performance, avoiding the need to deploy the same heavy-weight model for every problem, no matter how simple.

In the original release of the multi-temporal crop segmentation benchmark \cite{jakubik2023foundationmodelsgeneralistgeospatial}, the dataset comprised 3,854 chips derived from the 2022 U.S.\ CDL data. Using \brandname’s seamless data pipeline we enlarged the dataset to 14,061 chips and generated a newer 2024 version with 18,675 chips. Fine-tuning Prithvi-V2-300M on the expanded 2022 data yields a mIoU of 60.65 \% and accuracy of 83.02 \%, gains of +12 pp and +16 pp, respectively, which establish a new state of the art (see Figure ~\ref{fig:panel4}a and Supplementary Table \ref{table:new-us-cdl-exp}). The 2024 variant preserves the accuracy gain but delivers only a +6 pp increase in mIoU, probably because its labels were produced by downsampling the new 10m CDL data \cite{li2024cloudagriculture} to 30m, while the 2022 variant was generated directly at 30m resolution. The new 10m CDL also incorporates distinct compositing, sampling, and ancillary-data strategies that differ from the earlier 30m CDL workflow \cite{li2024cloudagriculture}. 

Taken together, our results underscore the versatility of \brandname as an end‑to‑end geospatial ML framework for operational solutions built on multispectral satellite imagery. \brandname\ was designed to serve three communities: (i) policymakers, who can generate on‑demand maps to guide decisions; (ii) engineers, who can query a REST API to integrate predictions into their systems; and (iii) researchers, who can ingest their own \emph{in situ} labels, create chips with \texttt{chip\_creator}, fine‑tune models, and deploy them locally or in the cloud. By releasing all data creation and training scripts alongside an interactive web application, we enable users to choose models for a downstream task, delineate regions of interest via a bounding box, and obtain predictions directly in the browser, promoting reproducibility, accessibility, and rapid deployment of geospatial AI solutions.

\section*{Methods}

\section*{\brandname model} \label{sec:model_methods}
\subsection*{Architecture}

InstaGeo is an end‑to‑end framework for geospatial machine learning that (i) curates geospatial datasets, (ii) derives lightweight task‑specific models, and (iii) deploys results to an interactive web-map. By “end‑to‑end” we mean that InstaGeo covers the entire workflow: data creation, GFM fine‑tuning, and operational deployment, without requiring the user to switch tools. Two common bottlenecks motivate InstaGeo’s design. First, geo‑experts often possess \emph{in‑situ} observations but lack the machine learning expertise needed to leverage GFMs. Second, ML practitioners may recognise the value of GFMs yet struggle with the intricacies of geospatial data handling. These gaps slow progress in Earth-observation research and limit the societal benefits of open‑source satellite imagery. InstaGeo bridges these gaps by providing three loosely coupled components, data, model, and application that transform raw geo-located observations into an operational system while allowing each expert to remain in their comfort zone. The remainder of this section details each component.

\subsubsection*{Data Component}
Subject matter experts collect \emph{in‑situ} observations as geolocated points or field polygons. To train or fine‑tune a GFM such as Prithvi‑EO‑2.0 \cite{szwarcman2025prithvieo20}, each observation must be paired with temporally matched satellite imagery (e.g. Sentinel‑2 or HLS). Constructing such datasets requires non-trivial geospatial expertise. InstaGeo automates this process through its \emph{chip\_creator} module, which produces multitemporal satellite image patches (called chips) of shape $T \times C \times H \times W$ ($T$ time steps, $C$ spectral bands, $H \times W$ spatial dimensions). The minimum inputs are longitude, latitude, date, and label (for points) or a label GeoTIFF and polygon geometry (for field polygons). Optional parameters such as number of timesteps, cloud-cover threshold, temporal tolerance, etc. allow further customization of the pipeline.

Once inputs are provided, the workflow proceeds as follows:

\begin{enumerate}
    \item \textbf{STAC Query}: For each observation, the module queries the STAC-compliant catalog\footnote{\url{https://stacspec.org/en}} of the selected satellite data product (currently HLS, Sentinel-2, Sentinel-1) for granules within each observation time window (determined by date, time-steps, step size, and tolerance).
    
    \item \textbf{Granule Filtering}: The retrieved metadata are filtered by cloud-cover threshold, valid coverage at the observation location, and, optionally, daytime acquisition.

    \item \textbf{Time-Step Assignment and Grouping}: For each observation, one granule is selected per time step. Observations that share identical granules in all $T$ time steps are grouped to maximize I/O efficiency.
    
    \item \textbf{Chunk Retrieval}: Each STAC granule, a Cloud-Optimized GeoTIFF, is opened, and the spatial window corresponding to the chip size (for points) or polygon bounds is read as the chip data. For point observations, multiple points in one chip are rasterized into a segmentation map; for polygon input, pre-existing segmentation maps are loaded.
    
    \item \textbf{Quality Control}: Cloud masking and other post-processing steps are applied. Pairs of chip and label are discarded if any valid label pixel lacks valid data at one or more time steps in the chip.
    
    \item \textbf{Serialisation}: At this step, we have a pair of valid chip ($T \times C \times H \times W$) and segmentation map ($H \times W$) that are written to disk as GeoTIFFs, ready for model training, fine-tuning or inference.
\end{enumerate}

Figure \ref{fig:panel1}a illustrates this pipeline. The resulting dataset seamlessly feeds into the model component for model training, fine-tuning or inference.

\subsubsection{Model Component}
\label{sec:model-component}
The model component ingests the chip–label pairs and produces a model for each dowsntream task by \emph{fine-tuning} or \emph{distilling} a GFM. We built on the Prithvi-EO-2.0 family, which natively accepts multitemporal inputs and has demonstrated strong downstream performance. Two training regimes are supported:

\begin{enumerate}
  \item \textbf{Vanilla fine-tuning}
  In this training regime, all weights of the pretrained GFM encoder and the attached segmentation head are updated at every training step. The resulting model retains the same architecture and parameter count, regardless of the dowsntream task.
  \item \textbf{Task-specific distillation}. 
  To tailor model complexity to each downstream task, we first perform vanilla fine-tuning to create a “teacher” model. We then initialise a “student” model using only the first $N$ encoder layers of the teacher, where $N$ is selected by experiments over values in \{2, 4, 6, …, $num\_encoder\_layers$\}. During distillation, the student’s parameters are updated to minimise the sum of the task loss (regression or classification loss) and a logits-level distillation loss against the frozen teacher (see Figure \ref{fig:panel1}b). This approach yields performance on par with vanilla fine-tuning while reducing both parameter count, and hence inference latency and $\mathrm{CO_{2}}$ footprint, by up to eightfold.
\end{enumerate}

In both regimes, we attach a task-agnostic segmentation head (transpose‐conv → conv → batch‐norm → dropout) that upsamples the encoded feature volume to an output of shape $C \times H \times W$, where $C$ is the number of classes (e.g., $C=1$ for regression, $C=\text{num\_classes}$ for segmentation).

\subsubsection{Application Component}
\label{sec: app-component}
The application component provides a unified full-stack geospatial analysis platform that guides users from data ingestion through model inference to interactive visualization. Its React-based single-page application lets users define areas of interest by drawing and editing bounding boxes on a web-map, with real-time area calculation and validation, and controls for model selection and inference parameter tuning. A FastAPI REST server handles task submission, status polling, and health checks, assigning each request a unique identifier and orchestrating a three-stage processing pipeline: first creating a chips dataset for the user-specified region, then running model inference on the dataset and finally preparing the chips and predictions to be served with TiTiler by converting them to Cloud Optimized GeoTIFFs (COGs). Once the pipeline is completed, the COGs outputs are rendered as interactive map layers, supporting overlays, so that end users can seamlessly explore both the satellite images used to run the inference and the model predictions and also export an analytic overview of the results to a PDF file. The system's microservices architecture also allows each component to be maintained and scaled independently to meet varying computational demands. 

\begin{figure}[hbt!]
\centering
\advance\leftskip-2cm
\captionsetup{width=1.0\linewidth}
\begin{center}
\includegraphics[width=1.0\textwidth]{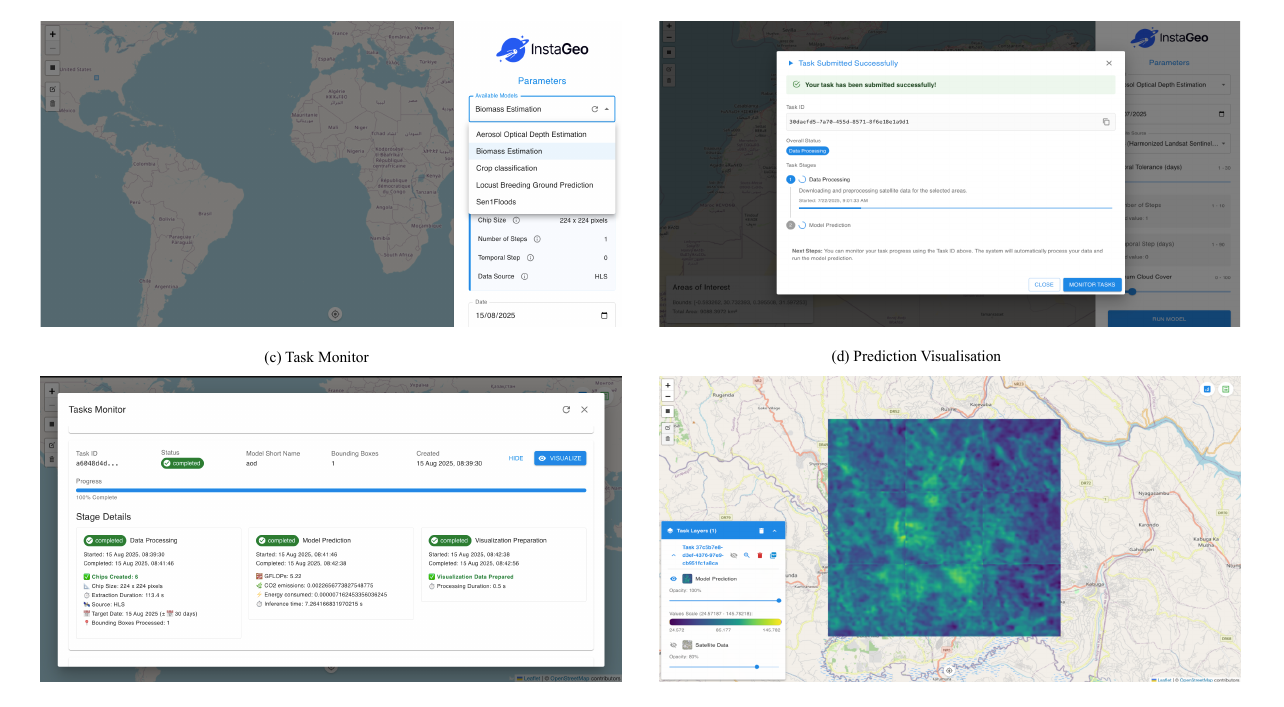}
\end{center}
\caption{\textbf{Lightweight web-map application - Application component.} (a) The user delineates a region of interest by drawing/editing bounding boxes and then selects the model, date, and other input parameters. (b) Submitting the request spawns data‑processing, inference and visualization preparation jobs that are queued for execution. (c) Job progress is tracked in the Tasks Monitor panel. (d) Upon completion, predictions are rendered interactively on the map.}
\label{fig:panelS2}
\end{figure}

The application component thus serves as a lightweight yet powerful web-map application that renders the GeoTIFF rasters produced in inference mode as interactive map layers (see Figure \ref{fig:panel1}b for the architecture and Figure \ref{fig:panelS2} shows the interface), while providing a complete end-to-end platform for geospatial analysis workflows.

\section*{Conclusion}
In this paper, we introduce InstaGeo, an end-to-end framework comprising three integrated components, aimed at enhancing and streamlining the transition from data acquisition to operational deployment in geospatial machine learning. With InstaGeo, users can efficiently generate training-ready datasets from in-situ labels and open-source satellite imagery, fine-tune geospatial foundation models on downstream tasks, and operationalize the system through an application that enables the selection of areas of interest, model inference, and the extraction of high-level insights from predictions. We demonstrate the utility of InstaGeo by reproducing both datasets and model results from three published benchmarks, thereby highlighting its reliability, reproducibility, and potential to accelerate research and applications in geospatial machine learning. As part of future releases, we plan to conduct new series of experiments, benchmark novel model architectures across downstream tasks not addressed in this paper, and extend the capabilities of the application component to further broaden its impact and usability.




\newpage
\bibliography{main}

\begin{thebibliography}{18}
\providecommand{\natexlab}[1]{#1}
\providecommand{\url}[1]{\texttt{#1}}
\expandafter\ifx\csname urlstyle\endcsname\relax
  \providecommand{\doi}[1]{doi: #1}\else
  \providecommand{\doi}{doi: \begingroup \urlstyle{rm}\Url}\fi

\bibitem[Roy et~al.(2014)Roy, Wulder, Loveland, Woodcock, Allen, Anderson, and Zhu]{roy2014landsat8}
D.~P. Roy, M.~A. Wulder, T.~R. Loveland, C.~E. Woodcock, R.~G. Allen, M.~C. Anderson, and Z.~Zhu.
\newblock Landsat‐8: Science and product vision for terrestrial global change research.
\newblock \emph{Remote Sensing of Environment}, 145:\penalty0 154--172, 2014.
\newblock \doi{10.1016/j.rse.2014.02.001}.

\bibitem[Masek et~al.(2020)Masek, Wulder, Markham, McCorkel, Crawford, Storey, and Jenstrom]{masek2020landsat9}
J.~G. Masek, M.~A. Wulder, B.~L. Markham, J.~McCorkel, C.~J. Crawford, J.~Storey, and D.~T. Jenstrom.
\newblock Landsat 9: Empowering open science and applications through continuity.
\newblock \emph{Remote Sensing of Environment}, 248:\penalty0 111968, 2020.
\newblock \doi{10.1016/j.rse.2020.111968}.

\bibitem[Drusch et~al.(2012)Drusch, Del~Bello, Carlier, Colin, Fernández, Gascon, and Bargellini]{drusch2012sentinel2}
M.~Drusch, U.~Del~Bello, S.~Carlier, O.~Colin, V.~M. Fernández, F.~Gascon, and P.~Bargellini.
\newblock Sentinel-2: Esa's optical high-resolution mission for gmes operational services.
\newblock \emph{Remote Sensing of Environment}, 120:\penalty0 25--36, 2012.
\newblock \doi{10.1016/j.rse.2011.11.026}.

\bibitem[Gascon et~al.(2014)Gascon, Martínez, and Heckelen]{gascon2014calval}
F.~Gascon, P.~Martínez, and C.~Heckelen.
\newblock Copernicus sentinel-2 mission: Products, algorithms and cal/val.
\newblock In \emph{Earth Observing Systems XVIII, Proceedings of SPIE}, volume 9218, 2014.
\newblock \doi{10.1117/12.2062260}.

\bibitem[Kansakar and Hossain(2016)]{kansakar2016review}
P.~Kansakar and F.~Hossain.
\newblock A review of applications of satellite earth observation data for global societal benefit and stewardship of planet earth.
\newblock \emph{Space Policy}, 36:\penalty0 46--54, 2016.
\newblock ISSN 0265-9646.
\newblock \doi{10.1016/j.spacepol.2016.05.005}.

\bibitem[Phiri et~al.(2020)Phiri, Simwanda, Salekin, Nyirenda, Murayama, and Ranagalage]{phiri2020sentinel2review}
D.~Phiri, M.~Simwanda, S.~Salekin, V.~R. Nyirenda, Y.~Murayama, and M.~Ranagalage.
\newblock Sentinel-2 data for land cover/use mapping: A review.
\newblock \emph{Remote Sensing}, 12\penalty0 (14):\penalty0 2291, 2020.
\newblock \doi{10.3390/rs12142291}.

\bibitem[Claverie et~al.(2018)Claverie, Ju, Masek, Dungan, Vermote, Roger, Skakun, and Justice]{HLS}
M.~Claverie, J.~Ju, J.G. Masek, J.L. Dungan, E.F. Vermote, J.-C. Roger, S.V. Skakun, and C.~Justice.
\newblock The harmonized landsat and sentinel-2 surface reflectance data set.
\newblock \emph{Remote Sensing of Environment}, 219:\penalty0 145--161, 2018.
\newblock \doi{10.1016/j.rse.2018.09.002}.
\newblock URL \url{https://www.sciencedirect.com/science/article/pii/S0034425718304139}.

\bibitem[Tuia et~al.(2025)Tuia, Schindler, Demir, Zhu, Kochupillai, Džeroski, van Rijn, Hoos, Del~Frate, Datcu, Markl, Le~Saux, Schneider, and Camps-Valls]{Tuia_2025}
D.~Tuia, K.~Schindler, B.~Demir, X.X. Zhu, M.~Kochupillai, S.~Džeroski, J.N. van Rijn, H.H. Hoos, F.~Del~Frate, M.~Datcu, V.~Markl, B.~Le~Saux, R.~Schneider, and G.~Camps-Valls.
\newblock Artificial intelligence to advance earth observation: A review of models, recent trends, and pathways forward.
\newblock \emph{IEEE Geoscience and Remote Sensing Magazine}, pages 2--25, 2025.
\newblock \doi{10.1109/MGRS.2024.3425961}.
\newblock URL \url{http://dx.doi.org/10.1109/MGRS.2024.3425961}.

\bibitem[Cong et~al.(2023)Cong, Khanna, Meng, Liu, Rozi, He, Burke, Lobell, and Ermon]{satmae}
Y.~Cong, S.~Khanna, C.~Meng, P.~Liu, E.~Rozi, Y.~He, M.~Burke, D.B. Lobell, and S.~Ermon.
\newblock Satmae: Pre-training transformers for temporal and multi-spectral satellite imagery.
\newblock \emph{arXiv preprint}, 2023.
\newblock URL \url{https://arxiv.org/abs/2207.08051}.
\newblock arXiv:2207.08051 [cs.CV].

\bibitem[Xiong et~al.(2024)Xiong, Wang, Zhang, Stewart, Hanna, Borth, Papoutsis, Le~Saux, Camps-Valls, and Zhu]{dofa}
Z.~Xiong, Y.~Wang, F.~Zhang, A.J. Stewart, J.~Hanna, D.~Borth, I.~Papoutsis, B.~Le~Saux, G.~Camps-Valls, and X.X. Zhu.
\newblock Neural plasticity-inspired multimodal foundation model for earth observation.
\newblock \emph{arXiv preprint}, 2024.
\newblock URL \url{https://arxiv.org/abs/2403.15356}.
\newblock arXiv:2403.15356 [cs.CV].

\bibitem[Mendieta et~al.(2023)Mendieta, Han, Shi, Zhu, and Chen]{gfm}
M.~Mendieta, B.~Han, X.~Shi, Y.~Zhu, and C.~Chen.
\newblock Towards geospatial foundation models via continual pretraining.
\newblock \emph{arXiv preprint}, 2023.
\newblock URL \url{https://arxiv.org/abs/2302.04476}.
\newblock arXiv:2302.04476 [cs.CV].

\bibitem[Jakubik et~al.(2023)Jakubik, Roy, Phillips, Fraccaro, Godwin, Zadrozny, Szwarcman, Gomes, Nyirjesy, Edwards, Kimura, Simumba, Chu, Mukkavilli, Lambhate, Das, Bangalore, Oliveira, Muszynski, Ankur, Ramasubramanian, Gurung, Khallaghi, Li, Cecil, Ahmadi, Kordi, Alemohammad, Maskey, Ganti, Weldemariam, and Ramachandran]{jakubik2023foundationmodelsgeneralistgeospatial}
Johannes Jakubik, Sujit Roy, C.~E. Phillips, Paolo Fraccaro, Denys Godwin, Bianca Zadrozny, Daniela Szwarcman, Carlos Gomes, Gabby Nyirjesy, Blair Edwards, Daiki Kimura, Naomi Simumba, Linsong Chu, S.~Karthik Mukkavilli, Devyani Lambhate, Kamal Das, Ranjini Bangalore, Dario Oliveira, Michal Muszynski, Kumar Ankur, Muthukumaran Ramasubramanian, Iksha Gurung, Sam Khallaghi, Hanxi Li, Michael Cecil, Maryam Ahmadi, Fatemeh Kordi, Hamed Alemohammad, Manil Maskey, Raghu Ganti, Kommy Weldemariam, and Rahul Ramachandran.
\newblock Foundation models for generalist geospatial artificial intelligence.
\newblock \emph{arXiv preprint arXiv:2310.18660}, 2023.
\newblock URL \url{https://arxiv.org/abs/2310.18660}.

\bibitem[Hong et~al.(2024)Hong, Zhang, Li, Li, Li, Yao, Yokoya, Li, Ghamisi, Jia, Plaza, Gamba, Benediktsson, and Chanussot]{spectral_gpt}
D.~Hong, B.~Zhang, X.~Li, Y.~Li, C.~Li, J.~Yao, N.~Yokoya, H.~Li, P.~Ghamisi, X.~Jia, A.~Plaza, P.~Gamba, J.A. Benediktsson, and J.~Chanussot.
\newblock Spectralgpt: Spectral remote sensing foundation model.
\newblock \emph{IEEE Transactions on Pattern Analysis and Machine Intelligence}, 46\penalty0 (8):\penalty0 5227--5244, 2024.
\newblock \doi{10.1109/TPAMI.2024.3362475}.
\newblock URL \url{http://dx.doi.org/10.1109/TPAMI.2024.3362475}.

\bibitem[Szwarcman et~al.(2025)Szwarcman, Roy, Fraccaro, Gíslason, Blumenstiel, Ghosal, de~Oliveira, Almeida, Sedona, Kang, Chakraborty, Wang, Gomes, Kumar, Truong, Godwin, Lee, Hsu, Asanjan, Mujeci, Shidham, Keenan, Arevalo, Li, Alemohammad, Olofsson, Hain, Kennedy, Zadrozny, Bell, Cavallaro, Watson, Maskey, Ramachandran, and Moreno]{szwarcman2025prithvieo20}
D.~Szwarcman, S.~Roy, P.~Fraccaro, Þ.E. Gíslason, B.~Blumenstiel, R.~Ghosal, P.H. de~Oliveira, J.L.S. Almeida, R.~Sedona, Y.~Kang, S.~Chakraborty, S.~Wang, C.~Gomes, A.~Kumar, M.~Truong, D.~Godwin, H.~Lee, C.-Y. Hsu, A.A. Asanjan, B.~Mujeci, D.~Shidham, T.~Keenan, P.~Arevalo, W.~Li, H.~Alemohammad, P.~Olofsson, C.~Hain, R.~Kennedy, B.~Zadrozny, D.~Bell, G.~Cavallaro, C.~Watson, M.~Maskey, R.~Ramachandran, and J.B. Moreno.
\newblock Prithvi-eo-2.0: A versatile multi-temporal foundation model for earth observation applications.
\newblock \emph{arXiv preprint}, 2025.

\bibitem[Spradlin et~al.(2024)Spradlin, Caraballo-Vega, Li, Carroll, Gong, and Montesano]{spradlin2024satvisiontoa}
C.S. Spradlin, J.A. Caraballo-Vega, J.~Li, M.L. Carroll, J.~Gong, and P.M. Montesano.
\newblock Satvision-toa: A geospatial foundation model for coarse-resolution all-sky remote sensing imagery.
\newblock \emph{arXiv preprint}, 2024.
\newblock URL \url{https://arxiv.org/abs/2411.17000}.
\newblock arXiv:2411.17000 [cs.CV].

\bibitem[Lacoste et~al.(2023)Lacoste, Lehmann, Rodriguez, Sherwin, Kerner, Lütjens, Irvin, Dao, Alemohammad, Drouin, Gunturkun, Huang, Vazquez, Newman, Bengio, Ermon, and Zhu]{lacoste2023geobench}
A.~Lacoste, N.~Lehmann, P.~Rodriguez, E.D. Sherwin, H.~Kerner, B.~Lütjens, J.A. Irvin, D.~Dao, H.~Alemohammad, A.~Drouin, M.~Gunturkun, G.~Huang, D.~Vazquez, D.~Newman, Y.~Bengio, S.~Ermon, and X.X. Zhu.
\newblock Geo-bench: Toward foundation models for earth monitoring.
\newblock \emph{arXiv preprint}, 2023.
\newblock URL \url{https://arxiv.org/abs/2306.03831}.
\newblock arXiv:2306.03831 [cs.LG].

\bibitem[Yusuf et~al.(2024)Yusuf, Yusuf, Panford-Quainoo, and Pretorius]{yusuf2024geospatialapproachpredictingdesert}
I.S. Yusuf, M.O. Yusuf, K.~Panford-Quainoo, and A.~Pretorius.
\newblock A geospatial approach to predicting desert locust breeding grounds in africa.
\newblock \emph{arXiv preprint}, 2024.
\newblock URL \url{https://arxiv.org/abs/2403.06860}.
\newblock arXiv:2403.06860 [cs.LG].

\bibitem[Li et~al.(2024)Li, Mueller, Yang, Johnson, and Willis]{li2024cloudagriculture}
Z.~Li, R.~Mueller, Z.~Yang, D.~Johnson, and P.~Willis.
\newblock Cloud-powered agricultural mapping: A revolution toward 10m resolution cropland data layers.
\newblock In \emph{IGARSS 2024 - IEEE International Geoscience and Remote Sensing Symposium}, pages 4081--4084, 2024.
\newblock \doi{10.1109/IGARSS53475.2024.10641079}.
\newblock URL \url{https://www.nass.usda.gov/Research_and_Science/Cropland/docs/IGARSS2024_Proceedings_10mCDL_Li_etal.pdf}.

\end{thebibliography}






\clearpage

\end{document}